\def\year{2021}\relax
\documentclass[letterpaper]{article} 
\usepackage{aaai21}  
\usepackage{times}  
\usepackage{helvet} 
\usepackage{courier}  
\usepackage[hyphens]{url}  
\usepackage{graphicx} 
\usepackage{natbib}  
\usepackage{caption} 
\usepackage[switch]{lineno} %
\usepackage{url}
\usepackage{algorithm} 
\usepackage{algpseudocode}
\usepackage{multirow}
\usepackage{subfig}
\usepackage{makecell}

\algrenewcommand\algorithmicensure{\textbf{Server executes:}}

\title{Adaptive Federated Dropout:\\ Improving Communication Efficiency and Generalization for Federated Learning}
\author {

        Nader Bouacida \textsuperscript{\rm 1},
        Jiahui Hou \textsuperscript{\rm 1},
        Hui Zang \textsuperscript{\rm 2},
        Xin Liu \textsuperscript{\rm 1} \\
}
\affiliations {
    \textsuperscript{\rm 1} University of California, Davis \\
    \textsuperscript{\rm 2} Google \\
    nbouacida@ucdavis.edu, jhhou.cs@gmail.com, huizang@gmail.com, xinliu@ucdavis.edu
}

\begin{document}

\maketitle

\begin{abstract}
With more regulations tackling users' privacy-sensitive data protection in recent years, access to such data has become increasingly restricted and controversial. To exploit the wealth of data generated and located at distributed entities such as mobile phones, a revolutionary decentralized machine learning setting, known as Federated Learning, enables multiple clients located at different geographical locations to collaboratively learn a machine learning model while keeping all their data on-device. However, the scale and decentralization of federated learning present new challenges. Communication between the clients and the server is considered a main bottleneck in the convergence time of federated learning.

In this paper, we propose and study Adaptive Federated Dropout (AFD), a novel technique to reduce the communication costs associated with federated learning. It optimizes both server-client communications and computation costs by allowing clients to train locally on a selected subset of the global model. We empirically show that this strategy, combined with existing compression methods, collectively provides up to 57$\times$ reduction in convergence time. It also outperforms the state-of-the-art solutions for communication efficiency. Furthermore, it improves model generalization by up to 1.7\%. 
\end{abstract}

\section{Introduction}

Equipping smart personal devices with sensors, combined with the fact they are frequently carried, means that they offer access to a large amount of training data necessary for building reliable models. In many scenarios, the raw data is privacy-sensitive, making it impractical to harvest into a central server. Federated learning~\cite{Konen2016,McMahan2017,Bonawitz2019,Yang2019} addressed these issues by enabling mobile devices to collaboratively learn a shared global model using their training data stored locally under the coordination of a central server, decoupling training deep learning models from the need to collect and store the data in the cloud. 

The participating devices (usually referred to as clients) are large in numbers and often connected to the server via slow or unstable internet connections. With increasingly improved hardware capabilities in personal computing devices and larger deep learning models, the computation burden becomes less significant, and the communication overhead constitutes the main bottleneck of federated learning~\cite{Jakub2016,Smith2018,Kairouz2019}. Together, a limited network bandwidth and a great number of clients exacerbate the communication bottlenecks, increasing both the number of stragglers and the dropping probability of clients with restricted bandwidth or limited connectivity. With hundreds of communications rounds eventually needed for the model to converge, communication time poses a particular challenge to federated learning.

A naive implementation of the federated learning framework requires that each client sends a full model update back to the server in each training round, which can lead to discarding the clients with limited bandwidth in the training process. In this case, the learned model is likely to result in a degraded user experience during the inference stage for clients with restricted bandwidth~\cite{Nishio2019}. Training low capacity models with smaller communication footprint comes at the expense of model performance. Using additional computation to decrease the number of rounds of communication needed to train a model has its limits in terms of the number of local epochs performed in local training~\cite{McMahan2017} and cannot solve the straggler problem. Lossy compression strategies~\cite{Jakub2016} can cause a loss of accuracy in the long run and may not be enough to tackle the communication time reduction on their own~\cite{Dutta2020}. Besides, it is unclear how these strategies deal with statistical challenges associated with decentralized highly personalized and heterogeneous data. As such, our goal is to expand federated learning capabilities by further enhancing the communication efficiency of the training process.  

In this work, we explore a novel strategy of training a carefully-chosen subset of the global model in each communication round to reduce the communication footprint of both download and upload links. Used jointly with compression techniques, we notice that selective dropping of a subset of the model would significantly reduce the number of weights that need to be exchanged with the server without degrading the global model quality. The specific contributions of this paper are as follows:

\begin{itemize}
 \item We introduce \emph{Adaptive Federated Dropout}, a method that builds upon the idea of \emph{Federated Dropout}~\cite{Caldas2018}. Our approach allows each client to train a sub-model dynamically selected in each round based on a score map of the model's activations while still providing updates applicable to the larger global model on the server-side. Thus, it reduces communication costs by exchanging these smaller sub-models between the clients and the server instead of the full model updates while also decreasing the computation by training on a subset of the model's weights.   
  \item We investigate the effect of our techniques on the convergence of the federated learning algorithms in terms of generalization and wall-clock time. We are able to significantly reduce the total amount of data transferred and communication delays, resulting in a speedup of up to 57$\times$ of the overall convergence time compared to up to 44$\times$ for the state-of-the-art, while improving the global model's generalization ability.
\end{itemize}

\section{Related Work}

Researchers have proposed several approaches to overcome the communication bottleneck in federated learning and distributed training in general. Authors in~\cite{McMahan2017} proposed the Federated Averaging algorithm, a practical method for the federated learning based on iterative model averaging. The main idea behind Federated Averaging is to compute higher quality updates rather than simple gradient steps~\cite{Hoffer2017}. It succeeded in training deep networks using 10-100x fewer communication rounds than a naively federated version of distributed Stochastic Gradient Decent (SGD). Similar efforts~\cite{Chen2016,Agarwal2018,Nilsson2018,Stich2019,Lin2020} introduced variants of local SGD, which perform several update steps on a local model before communicating with the server. These algorithmic solutions show performance gains in training efficiency and scalability to the underlying system resources. Nevertheless, downloading a large model from the server can still be a significant burden for clients, particularly those located in regions with limited connectivity.

Several works~\cite{Jakub2016,Lin2018,Wangni2018,Dutta2020} have proposed the compression of clients' updates to reduce the communication bandwidth, exploiting the redundancy observed in the gradient exchange in distributed SGD. Compression algorithms for federated learning can be put into two classes: The first category is the sketched updates, where clients compute a regular model update and perform a compression afterward. This class of methods includes subsampling, probabilistic quantization, and gradient sparsification. The second category of compression methods is the structured updates, where the model update is restricted to be of a form that allows for efficient compression during the optimization process. Low rank and random mask are two structures, which are frequently employed~\cite{Jakub2016}. To deal with the high sparsity in the model update, they also proposed several strategies for improving compression quality: momentum correction, local gradient accumulation and local gradient clipping for gradient sparsification methods~\cite{Lin2018}; applying basis transformations such Hadamard transformation and Kashin's representation for quantization methods~\cite{Lyubarskii2010}. These methods achieve high model compression ratios while nearly mitigating the problem of accuracy loss brought in by sparse updates. However, they were originally conceived for compressing the clients-to-server exchanges, where unbiased updates are averaged and eventually mitigate the information loss impact resulting from compression. Although the uplink is typically much slower than the downlink, the server-to-clients exchanges still pose a challenge to the server.

 The cited methods are either ingrained in the training process or must be applied after generating the trained model. Instead, Adaptive Federated Dropout dynamically selects sub-models extracted from the global model before sending them for the clients and the subsequent local training.
 
 \begin{figure*}[ht]
	\centering
	\includegraphics[width=\textwidth]{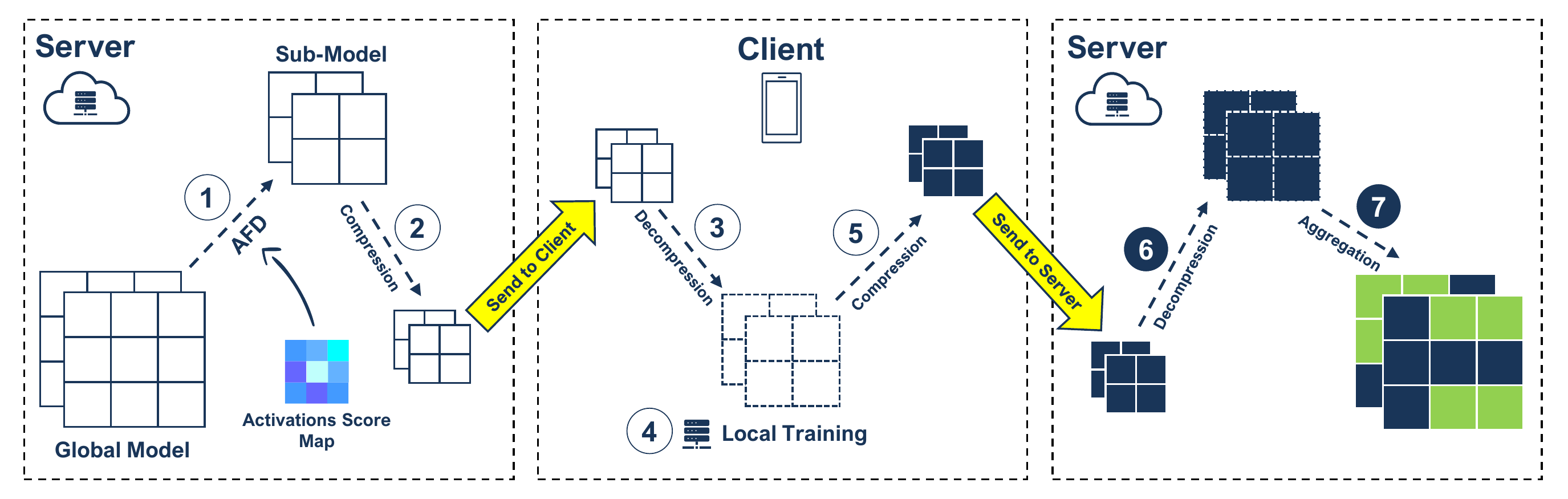}
	\caption{Adaptive Federated Dropout Overview -- We reduce the size of the models exchanged between the clients and the server by (1) building a sub-model using the activations score map, and (2) compressing the resulting structure. This compressed sub-model is sent to the client, which in turn (3) decompress it and (4) runs local training using its data, and (5) compress the updated sub-model afterward. The latter is sent back to the server, where (6) it will be decompressed, (7) recovered in its original shape, and finally aggregated with other updated models into the global model.}
	\label{afd_graph}
\end{figure*}

\section{Problem Statement}
In this paper, we mainly focus on synchronous federated learning algorithms which proceeds in rounds of training. It aims to learn a global model with parameters embodied in a real tensor $\mathcal{W}$ from data stored across a number of clients. In the training round $t \geq 1$, the server distributes the current global model $\mathcal{W}_t$ to the set of selected clients $\mathcal{S}_t$ with a total of $n_t$ data instances. The selected clients locally execute SGD based on their data and independently update the model. As the result, they produce local updated models $\{\mathcal{W}_t^c \mid c\in \mathcal{S}_t \}$. The update of client $c$ is expressed as:
\begin{equation}
    \mathcal{W}_t^c = \mathcal{W}_t - \alpha \mathcal{H}_t^c \qquad \forall c \in \mathcal{S}_t
\end{equation}
where $\mathcal{H}_t^c$ is the gradients tensor for client $c$ in training round~$t$ and $\alpha$ is the learning rate. Each selected client $c$ then sends the update back to the server. The server aggregate updates from all participating clients to construct the new global model $\mathcal{W}_{t+1}$ as follows: 
\begin{equation}
    \mathcal{W}_{t+1} = \frac{1}{n_t} \sum_{c \in \mathcal{S}_t} n_c \mathcal{W}_t^c
\end{equation}
where $n_c$ is the number of data instances of client $c$ and $n_t = \sum_{c \in \mathcal{S}_t} n_c$. Hence, we deduce that $\mathcal{W}_{t+1}$ can be written as:
\begin{equation}
    \mathcal{W}_{t+1} = \mathcal{W}_t - \alpha_t \mathcal{H}_t
\end{equation}
where $\mathcal{H}_t = \frac{1}{n_t} \sum_{c \in \mathcal{S}_t} n_c \mathcal{H}_t^c$.

The goal of improving the communication efficiency of federated learning is to reduce the cost of transferring model updates $\{\mathcal{W}_t^c \mid c\in \mathcal{S}_t \}$ from the clients to the server and the cost of transferring the global model $\mathcal{W}_t$ from the server to the clients. Our objective is to dynamically select a sub-model $w_t^{c}$ that reduces the convergence time without degrading the performance of the final model. In this work, our emphasis is on the critical nature of the communication constraints. Other issues related to federated learning networking problems, such as security or client availability, are beyond the scope of this work.

\section{Adaptive Federated Dropout}

Our proposed method named Adaptive Federated Dropout builds upon the technique Federated Dropout~\cite{Caldas2018}, yet is more flexible and efficient in reducing communications costs. Federated Dropout is inspired by the well-known regularisation strategy of dropout~\cite{Srivastava2014}. The main idea behind Federated Dropout is that each client trains a local update using a sub-model rather than training an update using the whole global model. The sub-models are subsets of the global model constructed by dropping a fixed percentage of the filters from convolutional layers and activations of fully-connected layers. While this method successfully decreases communication costs and local computation, it drops the activations randomly and does not consider the decomposition of the neural networks. It treats them as black-box functions without inspecting the changes in their internal structures resulting from dropping. Moreover, Federated Dropout was only conceived for conventional neural networks (CNN), and it is unclear how it can be extended to recurrent neural networks (RNN), the most popular types of models employed in natural language processing (NLP) tasks. 

Instead of randomly dropping a fraction of neurons, Adaptive Federated Dropout maintains an activation score map (see Figure~\ref{afd_graph}) to determine which activations should be selected to be transferred or dropped. Each score map assigns all the activations real values representing their importance and influence on the training process. In convolutional layers, dropping activations would not save any space, so we instead drop out a fixed percentage of the filters. Besides, the dropout is only applied to the non-recurrent connections for RNN models to preserve their memorization ability~\cite{Zaremba2014}. The server creates different reduced architectures in every training round based on the activations score map, meaning only the necessary parameters that are not affected by the selective dropping of the activations are transmitted. Since different neurons are discarded each round, each pass effectively calculates a gradient for a different sub-model. Following local training, the clients (which can be entirely unaware of the global model's architecture) send back the updated sub-model to the server. We propose two modes of Adaptive Federated Dropout: \emph{Multi-Model Adaptive Federated Dropout} and \emph{Single-Model Adaptive Federated Dropout}.

Next, we discuss the motivation of our proposed methods. First,  considering the capacity on the edge devices, small local datasets in the clients do not need to be trained on the full global model to achieve good generalization. 
There is substantial evidence that on-device personalization of model parameters has yielded significant improvements in the model performance~\cite{Wang2019, Zantedeschi2020}. While our goal is not personalization by any means, training a different customized compact sub-model for each client or cluster of clients will likely provide higher quality updates for the global model, especially with non-IID data distribution. This is observed in our experiment that our approach achieves better generalization. Furthermore, it is common that some compression methods~\cite{Lin2018, Zeyi2018} only transmit important gradients after training. We build and transmit sub-models constructed from only important activations. The difference is as follows. Gradient dropping does not create any specific model patterns and still force all the clients to train on the same shared global model. In comparison, dropping neurons, instead of gradients, results in different sub-model architectures before the training. 
These observations lead to our method's core idea: Different clients or clusters of clients are likely to use the sub-models differently, especially with high statistical heterogeneity. Training sub-models customized for the clients' data is likely to improve generalization. Apart from potential regularization effects, selecting these sub-models at random would not leverage the full benefits of Federated Dropout since different sub-models are not made equal with respect to generalization. 

\subsection{Multi-Model Adaptive Federated Dropout}

This strategy enables the server to keep a different activations score map for each client. At each round, the server sends a different sub-model for each selected client and updates each client's corresponding score map based on the local loss function. The main challenge of our work is how to select a group of important activations (Step 1 in Figure~\ref{afd_graph}) so that the training converges fast without accuracy loss. 

\begin{algorithm}[ht] 
\caption{Multi-Model Adaptive Federated Dropout}
\label{alg:dm_afd}
\textbf{Input:} Federated Dropout Rate $k\%$
\begin{algorithmic}[1]
\Ensure
\State \textbf{Initialize:} Activations score maps $\mathcal{M}_0, \dots, \mathcal{M}_n \leftarrow 0$ for the $n$ clients, Latest loss values for the $n$ clients $l_0, \dots, l_n \leftarrow 0$, $Recorded \leftarrow \textsc{false}$ 
\For {each round  $t = 1, 2, \dots, T$}
\State $\mathcal{S}_t \leftarrow $ (random set of $m$ clients) \Comment{$m \leq n$}
\For {each client  $c \in \mathcal{S}_t$ \textbf{in parallel}}
\If{$t > 1$}
  \If{$Recorded$}
  \State Select $A_c$ to create the sub-model $w_t^{c}$
    \Else
  \State $w_t^{c} \leftarrow$ Weighted Random Selection $k\%$
    \EndIf
\Else
\State $w_t^{c} \leftarrow$ Random Selection $k\%$
\EndIf
\State \textbf{After client $c$ executes local training on $w_t^{c}$:}
\State \hskip 2em $l_{t}^{c} \Leftarrow \ell(w_t^{c},c)$ \Comment{Local training loss}

\If{$l_{t}^{c} < l_c$}
  \State Record the indexes $A_c$ of sub-model $w_t^{c}$
  \State Update $\mathcal{M}_c$ at given activations indexes $A_c$
  \State $Recorded \leftarrow \textsc{true}$
\Else
  \State $Recorded \leftarrow \textsc{false}$
\EndIf

\State $l_c \leftarrow l_{t}^{c}$

\EndFor
\EndFor

\end{algorithmic}
\end{algorithm}

We introduce Multi-Model Adaptive Federated Dropout in Algorithm~\ref{alg:dm_afd}. The only parameter of AFD is the Federated Dropout Rate (FDR) $k$, representing the percentage of the activations to be dropped when constructing the sub-models. This parameter should be set empirically. We define $l_t^c$ as the loss function of client $c$ at round $t$, which we want to minimize. $l_c$ is a variable that saves the loss previously computed for client $c$. Besides, we initialize our score map $\mathcal{M}_c$ in each client $c$ with zeros (line 1) and randomly construct sub-models for the clients in the first training round (line 12).

Ideally, we want the loss function to get close to its optimal local value at each step. Nevertheless, the loss value usually fluctuates with the SGD optimizer.  Our technique tracks the loss values generated by the current local training round and the previous one (lines 15-16). If the current loss value $l_t^c$ is strictly lower than the previous loss saved in $l_c$, that indicates that we have chosen a better sub-model at round $t$. Hence, we record the indexes of the sub-model activations (line 17) and label them as important by signing a positive value equal to $\frac{l_c - l_t^c}{l_c}$ to their corresponding entries in $\mathcal{M}_c$ (line 18). For the subsequent round of local training, we use the same subset of activations $A_c$ (line 7) because these activations are proven beneficial to our loss function. 

Once we have a greater loss than previously found for the client $c$, that implies that at time $t$, the loss function value is not desirable. Therefore, we refrain from dropping the same model activations as the previous step. Instead, we randomly select the sub-model $w_t^c$ according to the activations score values (weighted random selection of the activations using weights from $\mathcal{M}_c$ -- line 9). We can think of the activations score map as a tensor that contains the scores describing how useful each activation. The lower an activation score is, the higher the chance this activation is getting dropped and vice versa. Indeed, the activations with great scores indicate that they are flagged as influential many times during the training process. Over time, different sub-models will be evaluated, and the score maps will reflect accurate estimates of the activations' importance. The weighted random selection based on the activations score map will likely generate the sub-model that best fits each client.   

\begin{algorithm}[t]
\caption{Single-Model Adaptive Federated Dropout}
\label{alg:sm_afd}
\textbf{Input:} Federated Dropout Rate $k\%$
\begin{algorithmic}[1]
\Ensure
\State \textbf{Initialize:} Activations score map $\mathcal{M} \leftarrow 0$, Latest average loss value $l \leftarrow 0$, , $Recorded \leftarrow \textsc{false}$ 
\For {each round  $t = 1, 2, \dots, T$}
\If{$t > 1$}
  \If{$Recorded$}
  \State Select $A$ to create the sub-model $w_{t}$
    \Else
  \State $w_{t} \leftarrow$ Weighted Random Selection $k\%$ 
    \EndIf
\Else
\State $w_t \leftarrow$ Random Selection $k\%$ 
\EndIf
\State $\mathcal{S}_t \leftarrow $ (random set of $m$ clients) \Comment{$m \leq N$}
\For {each client  $c \in \mathcal{S}_t$ \textbf{in parallel}}
\State \textbf{Client $c$ executes local training on $w_t$:}
\State \hskip 2em $l_{t}^{c} \Leftarrow \ell(w_t,c)$ \Comment{Local training loss}
\EndFor

\State $\bar{l}_t = \frac{1}{m} \sum_{c \in \mathcal{S}_t} l_{t}^{c}$ \Comment{Average training loss}

\If{$\bar{l}_t < l$}
  \State Record the indexes $A$ of sub-model $w_t$
  \State Update $\mathcal{M}$ at given activations indexes $A$
  \State $Recorded \leftarrow \textsc{true}$
\Else
  \State $Recorded \leftarrow \textsc{false}$
\EndIf
\State $l \leftarrow \bar{l}_t$
\EndFor

\end{algorithmic}
\end{algorithm}

\subsection{Single-Model Adaptive Federated Dropout}

The only difficulty with Multi-Model AFD is that training with a small fraction of clients at each round makes the algorithm behaves randomly, just like the standard Federated Dropout. We recall that Multi-Model AFD keeps an independent activations score map for each client. If the fraction of clients per round is small, each client will be elected few times for training, and thus its activations score map is not updated frequently enough to reflect accurate scores of the activations' importance.

To address this challenge, we investigate a second variant of AFD that creates the same sub-model $w_t$ for all selected clients $\mathcal{S}_t$ in each round $t$. As described in Algorithm~\ref{alg:sm_afd}, in this scenario, we keep only a single activations score map $\mathcal{M}$ at the server (line 1), and a single sub-model will be created based on this score map and distributed to all clients at a particular training round. This single score map $\mathcal{M}$ will be updated in a similar fashion to Multi-Model AFD (lines 18-24). However, we use the consecutive average losses $l$ and $\bar{l}_t$ of the selected clients from two subsequent rounds of training to update its values instead of the local losses (line 17). Moreover, the construction of the single sub-model follows the same logic as previously (lines 3-11) and we use $A$ to save the indexes of the sub-model $w_t$ in case that $\bar{l}_t < l$. This mode is more lightweight than the previous one. 


\section{Experimental Results}

\subsection{Experimental Setup}

We test our strategies using the already established federated learning benchmarks. In particular, we run all our experiments using the well-known Federated Averaging algorithm (FedAvg)~\cite{McMahan2017}.

\subsubsection{Datasets}
We conduct experiments on three different LEAF datasets~\cite{Caldas2018leaf}, a benchmark for federated settings. (1) FEMNIST dataset~\cite{Cohen2017} for 62-class image classification, which serves as a more complex extended version of the popular MNIST dataset~\cite{Lecun2010}. The data is partitioned based on the writer of the character in the non-IID setting. (2) Shakespeare dataset for next-character prediction, which is constructed from The Complete Works of William Shakespeare~\cite{Shakespeare1996}. Each role in each play is considered as a different client in the non-IID setting. (3) Sentiment140 dataset~\cite{Go2009} for 2-class sentiment analysis, which is generated by interpreting tweets based on the emoticons presented in them. Each twitter user is a client in the non-IID setting. In the IID setting, the data is sampled and randomly distributed over the clients. Thus, all users will have the same underlying distribution of data. We reserve 20\% of the data in each client for testing purposes.

\subsubsection{Models} For FEMNIST's image classification task, we use a CNN with two 5x5 convolution layers (the first one has 32 channels, the second one has 64 channels, each of them followed with 2 $\times$ 2 max-pooling), a fully connected dense layer with 2048 units, and a final softmax output layer. For Shakespeare dataset, we consider a two-layer LSTM classifier containing 256 hidden units preceded by an 8-dimensional embedding layer. The embedding layer takes a sequence of 80 characters as input, and the output is a class label between 0 and 52. Finally, for Sentiment140, the model used in the experiments is a two-layer LSTM classifier with 100 hidden units and pre-trained 300-dimensional GloVe embeddings~\cite{Pennington2014}. The input is a sequence of 25 words, where each word is embedded into a 300-dimensional space by looking up GloVe. The output of the last dense-connected layer is a binary classifier.
 
\subsubsection{Baselines} We optimize our experiments to work well in the baseline scenario, which does not involve any kind of Federated Dropout or compression. For local training at each client, we run a grid search on the learning rates, and we use the best-recorded learning rates: 0.004 for FEMNIST, 0.08 for Shakespeare, and 0.001 for Sentiment140. Each selected client trains for one local epoch per round using a batch size of 10.  We compare Adaptive Federated Dropout (AFD) to Federated Dropout (FD) and Deep Gradient Compression (DGC)~\cite{Lin2018}. The latter is one of the state-of-the-art compression methods for distributed training that employs gradient sparsification, momentum correction, and local gradient accumulation. 

\subsection{Results and Analysis}

\begin{figure*}[t] 
	\centering
	\subfloat[\emph{FEMNIST Dataset} \label{f1}]{\includegraphics[width=0.33\textwidth]{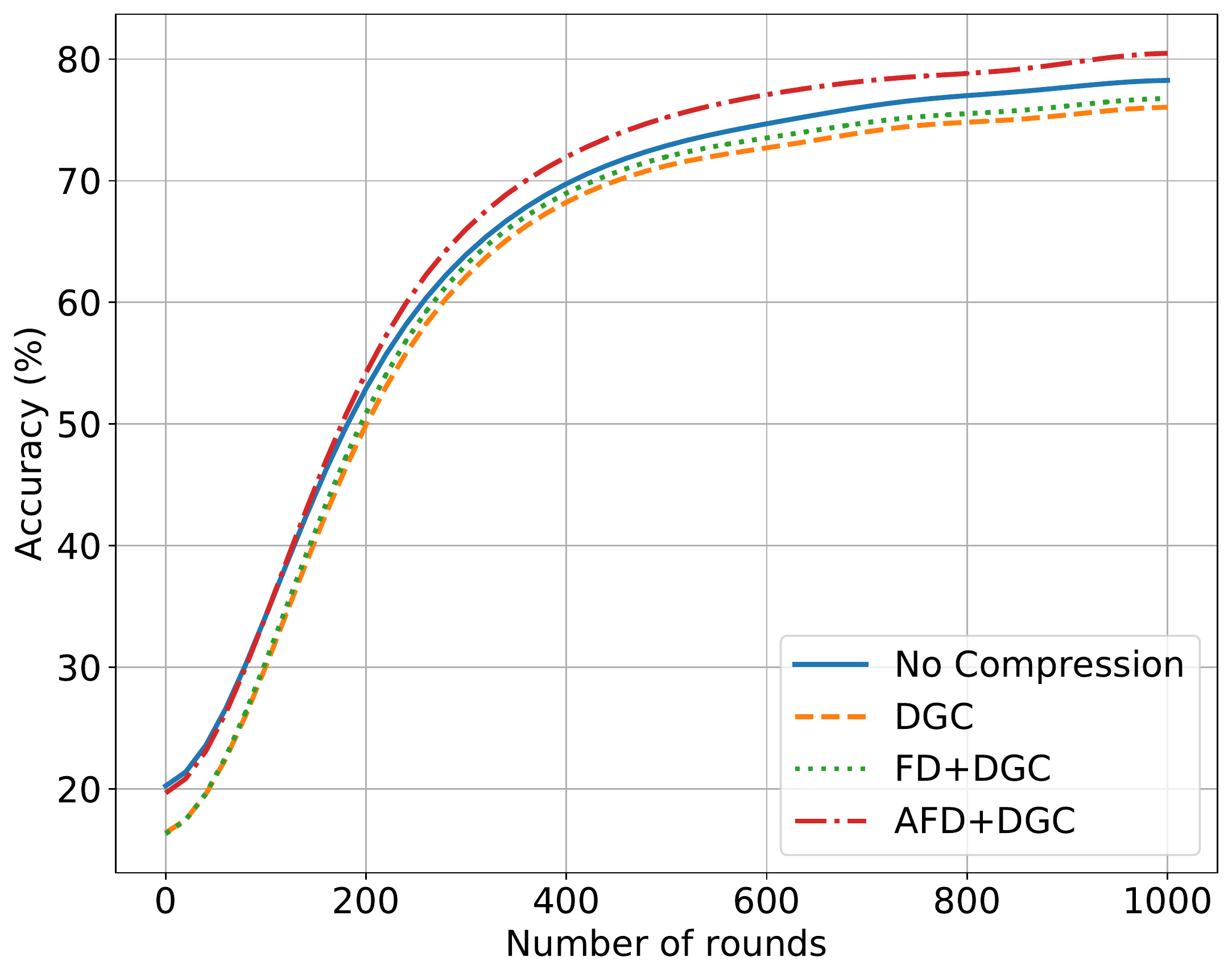}}\hfill
	\subfloat[\emph{Shakespeare Dataset} \label{sh1}]{\includegraphics[width=0.33\textwidth]{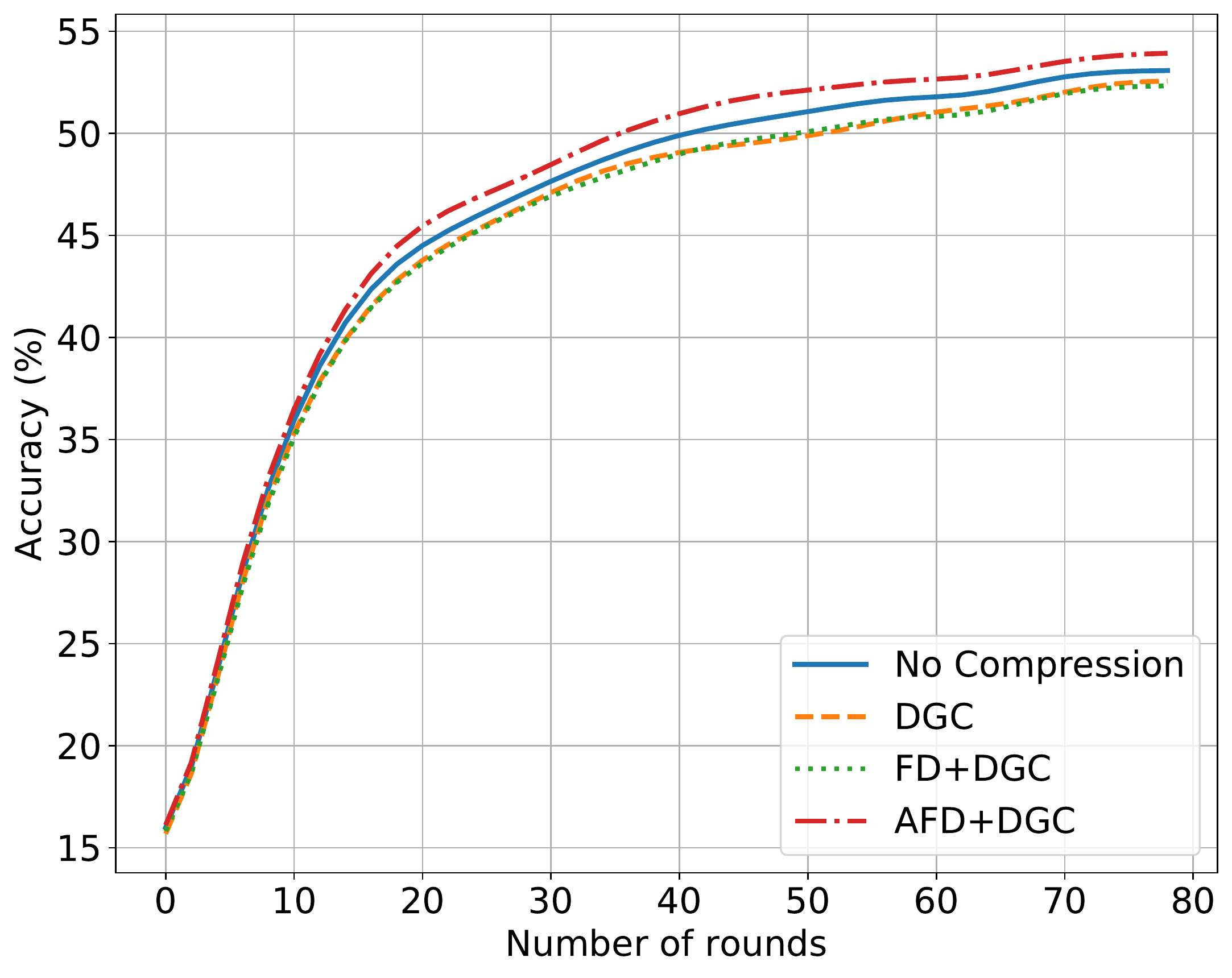}}\hfill
	\subfloat[\emph{Sentiment140 Dataset} \label{se1}]{\includegraphics[width=0.33\textwidth]{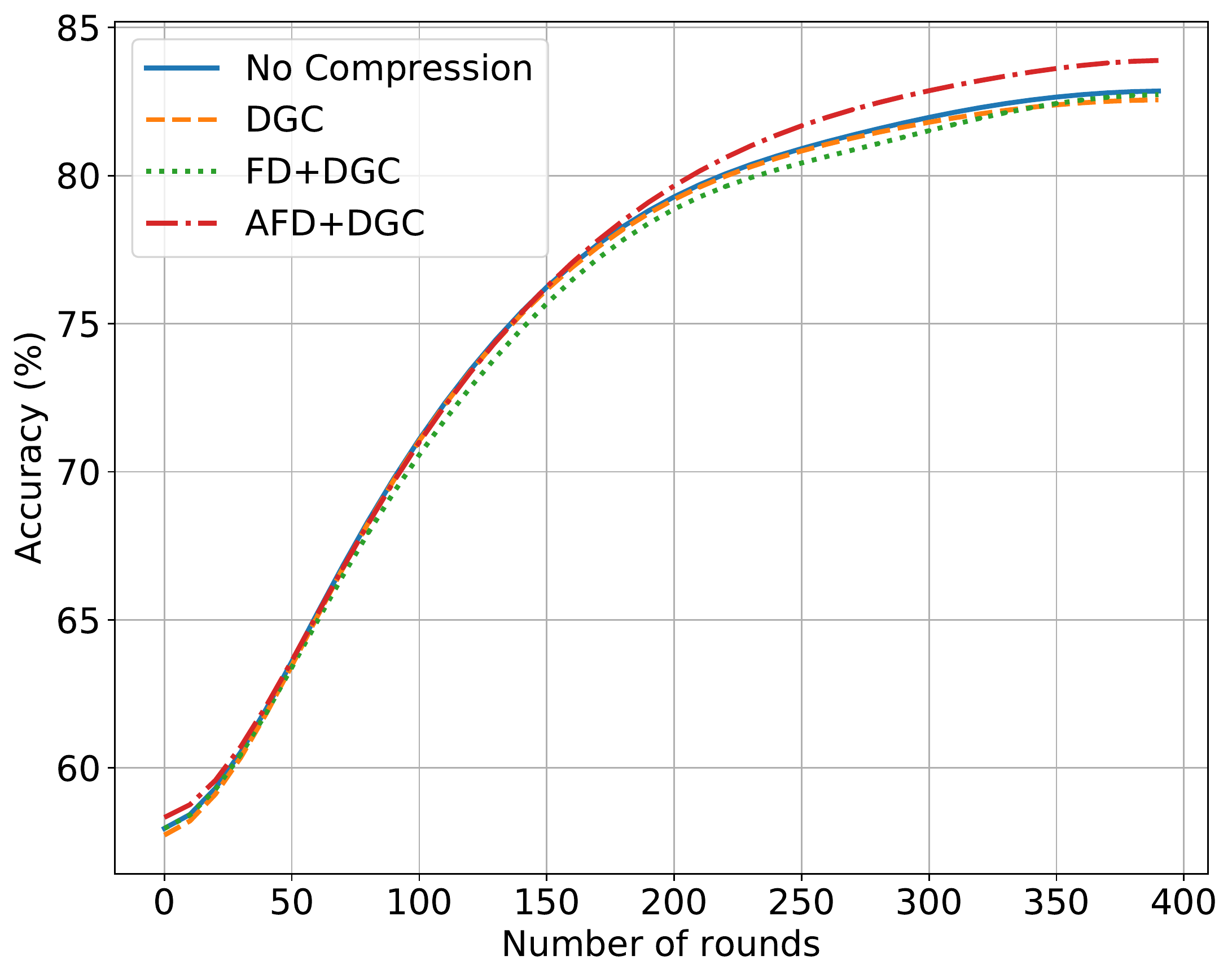}}\hfill
	\caption{Top-1 accuracy results for the non-IID version of the datasets and using Multi-Model AFD. }
	\label{fig2}
\end{figure*}

\begin{table*}[t]
\centering
\caption{Accuracy and convergence time results on the non-IID LEAF datasets. For FEMNIST, Shakespeare and Sentiment140 datasets, the models are trained for 1000, 80 and 400 rounds and, the target accuracy is set to 75\%, 50\% and 82\% respectively.}
\label{table1}
\begin{tabular}{c|l|c|c|c}
\Xhline{1pt}
\multicolumn{2}{l|}{}  &  Accuracy & Convergence Time (min) & Speedup Ratio \\ \hline
\multirow{4}{*}{FEMNIST}  & No Compression & 78.9\% $\pm$ 0.12\% &  3233.2 & 1$\times$ \\ 
                   & DGC & 76.3\% $\pm$ 0.43\% &  102.4 & 31$\times$ \\ 
                   & FD + DGC  & 77.5\% $\pm$ 0.24\% &  82.3 & 39$\times$ \\ 
                   & AFD + DGC  & \textbf{80.6\% $\pm$ 0.14\%} &  \textbf{61.7} & \textbf{52$\times$} \\ \hline
\multirow{4}{*}{Shakespeare}  & No Compression  & 53.1\% $\pm$ 0.22\% & 762.5 & 1$\times$ \\ 
                   & DGC & 52.8\% $\pm$ 0.54\% & 21.2 & 36$\times$ \\ 
                   & FD + DGC  & 52.5\% $\pm$ 0.34\% & 17.4 & 44$\times$ \\ 
                   & AFD + DGC  & \textbf{54.4\% $\pm$ 0.36\%} & \textbf{13.3} & \textbf{57$\times$} \\ \hline
\multirow{4}{*}{Sentiment140}  & No Compression  & 82.9\% $\pm$ 0.19\% & 3050.7 & 1$\times$ \\ 
                   & DGC & 82.5\% $\pm$ 0.29\% & 89.7 & 34$\times$ \\ 
                   & FD + DGC  & 82.7\% $\pm$ 0.11\% & 76.2 & 40$\times$ \\ 
                   & AFD + DGC  & \textbf{83.8\% $\pm$ 0.56\%} & \textbf{57.5} & \textbf{53$\times$} \\ 
\Xhline{1pt}
\end{tabular}
\end{table*}

\begin{figure*}[ht] 
	\centering
	\subfloat[\emph{FEMNIST Dataset} \label{f2}]{\includegraphics[width=0.33\textwidth]{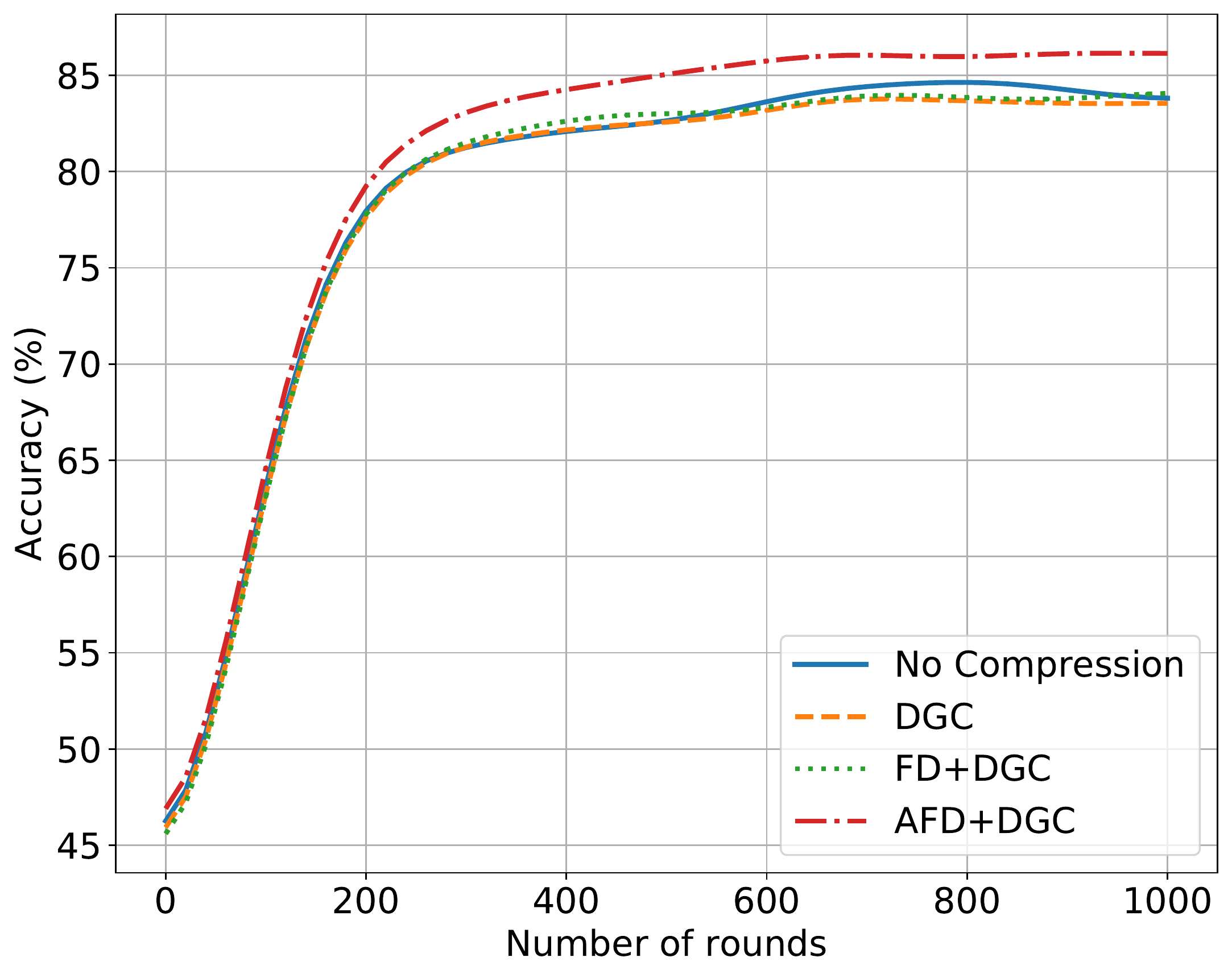}}\hfill
	\subfloat[\emph{Shakespeare Dataset} \label{sh}]{\includegraphics[width=0.33\textwidth]{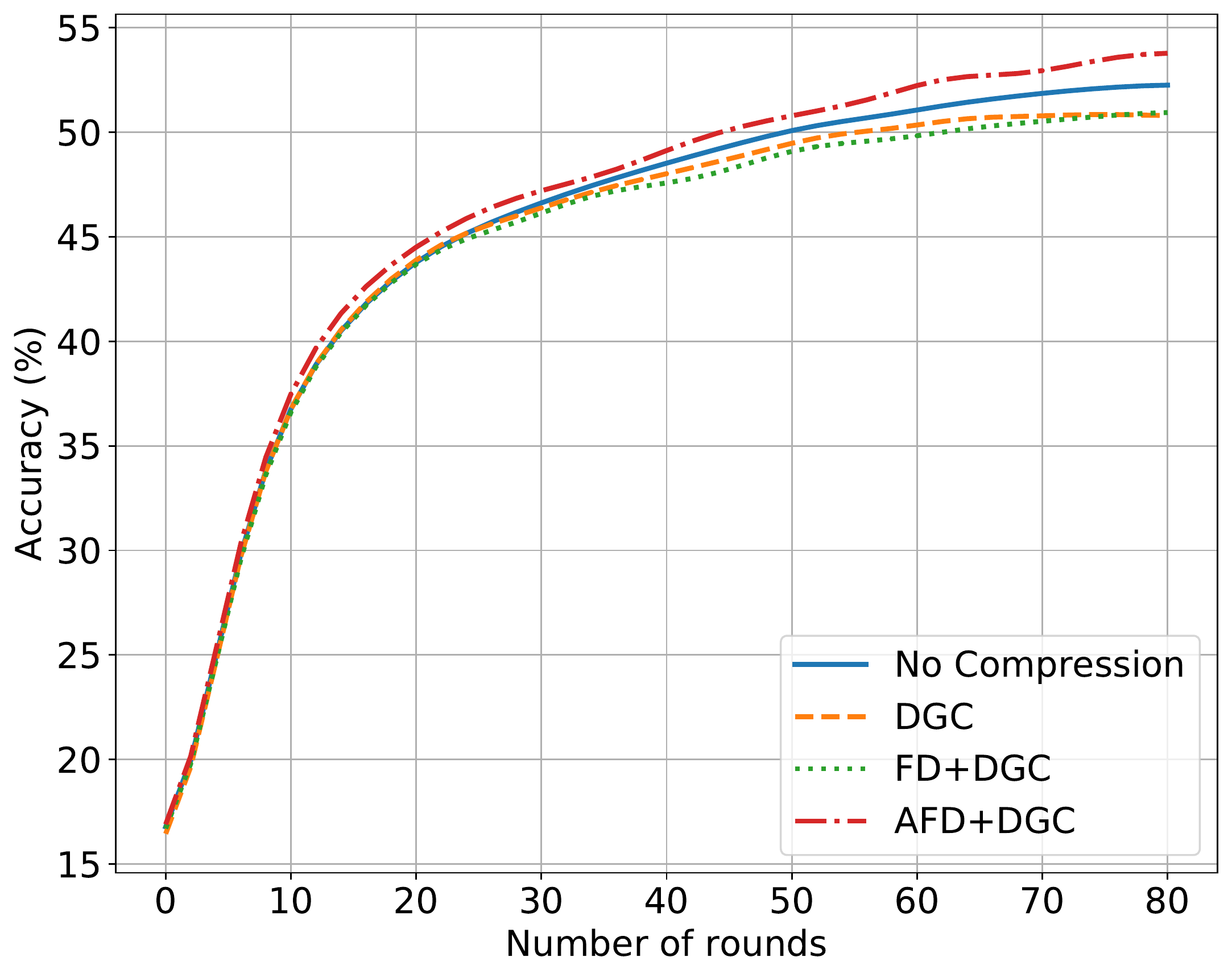}}\hfill
	\subfloat[\emph{Sentiment140 Dataset} \label{se2}]{\includegraphics[width=0.33\textwidth]{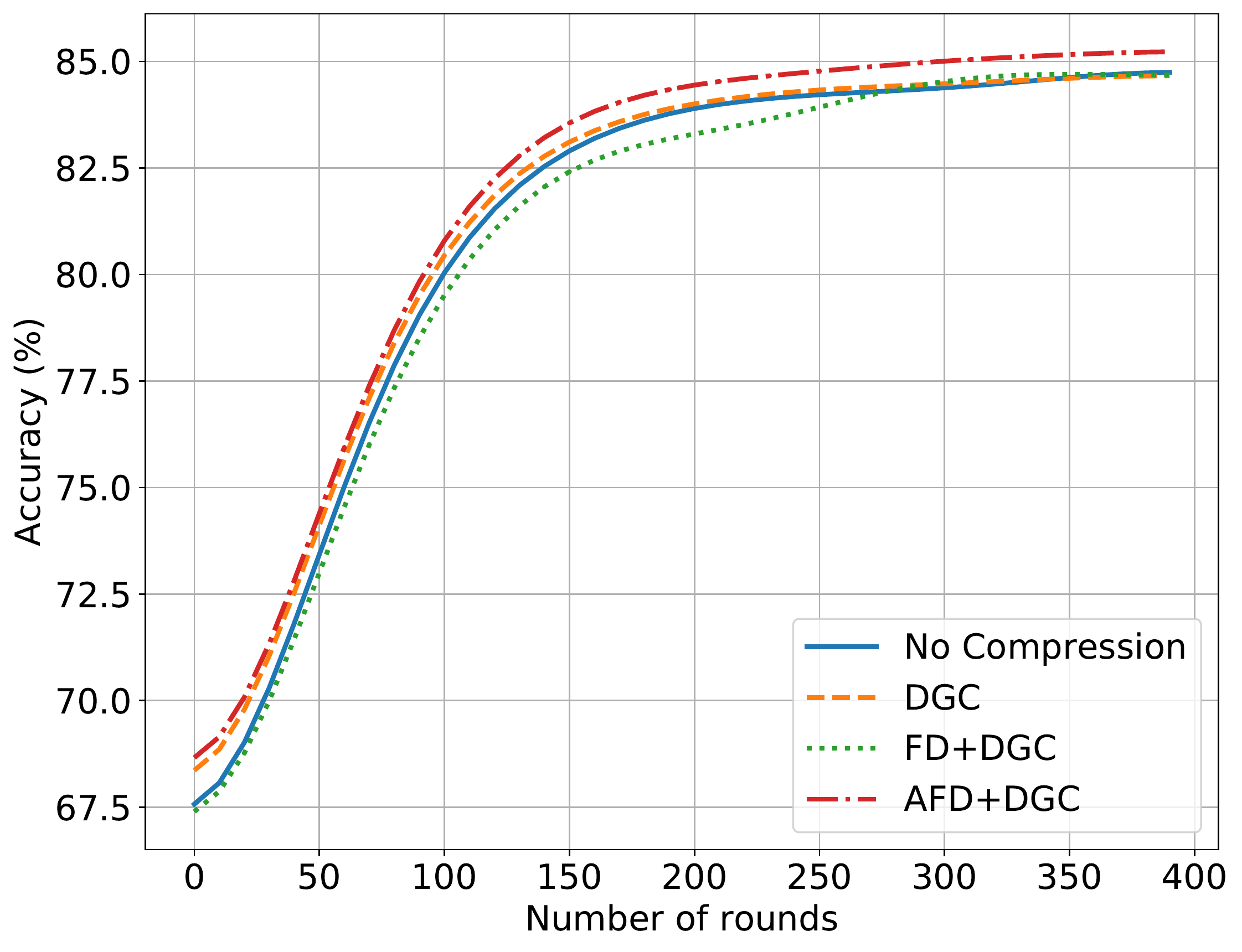}}\hfill
	\caption{Top-1 accuracy results for the IID version of the datasets and using Single-Model AFD. }
	\label{fig3}
\end{figure*}

We first examine Multi-Model AFD strategy. Figure~\ref{fig2} shows how the convergence (measured by the Top-1 accuracy of the global model) of our three applications behaves under different compression schemes when we simulate the non-IID version of the datasets. In these experiments, we compress all server-to-clients exchanges using 8-bit Gradient Quantization after applying Hadamard transformation as a basis function to spread the information on the compressed weights. We do not compress biases for any of the models because compressing smaller variables causes significant accuracy degradation but translates into minimal communications savings. Moreover,  we always keep the input and the output layers intact. We set the Federated Dropout Rate (FDR) as 25\%, and we randomly select 30\% of the clients for training in each round. The FDR parameter should be set empirically between 10\% and 50\%, taking into consideration the scale of the model. The higher FDR values are often possible with larger models. We combine both Adaptive Federated Dropout (AFD) and Federated Dropout (FD) with Deep Gradient Compression (DGC). We repeat each experiment 5 times with different seeds and report the mean among these repetitions. We note that DGC only operates on client-to-server communications because it is ingrained in the local training process.

The main takeaway from these experiments is that, for every model, Multi-Model AFD outperforms other methods. It seems to work across the board, not only preserving the global model quality but also achieving better accuracy. The selective dropping of the global model activations generates a compact sub-model that is more performant and efficient in learning from the client data. The fact that the accuracy of FD (with DGC compression) is very close to DGC means that our gains are not caused by some regularization effects. Indeed, the models employed are not large in size and, hence, FD has a tight margin to improve the generalization ability of the sub-model using regularization. Because some sub-models tend to generalize better than others for different clients, AFD is able to build the sub-models that best match each client's data, resulting in accuracy increases between 0.9\% and 1.7\% compared to the scenarios with no compression involved. 

\begin{figure}[ht]
\centering
\includegraphics[width=0.8\columnwidth]{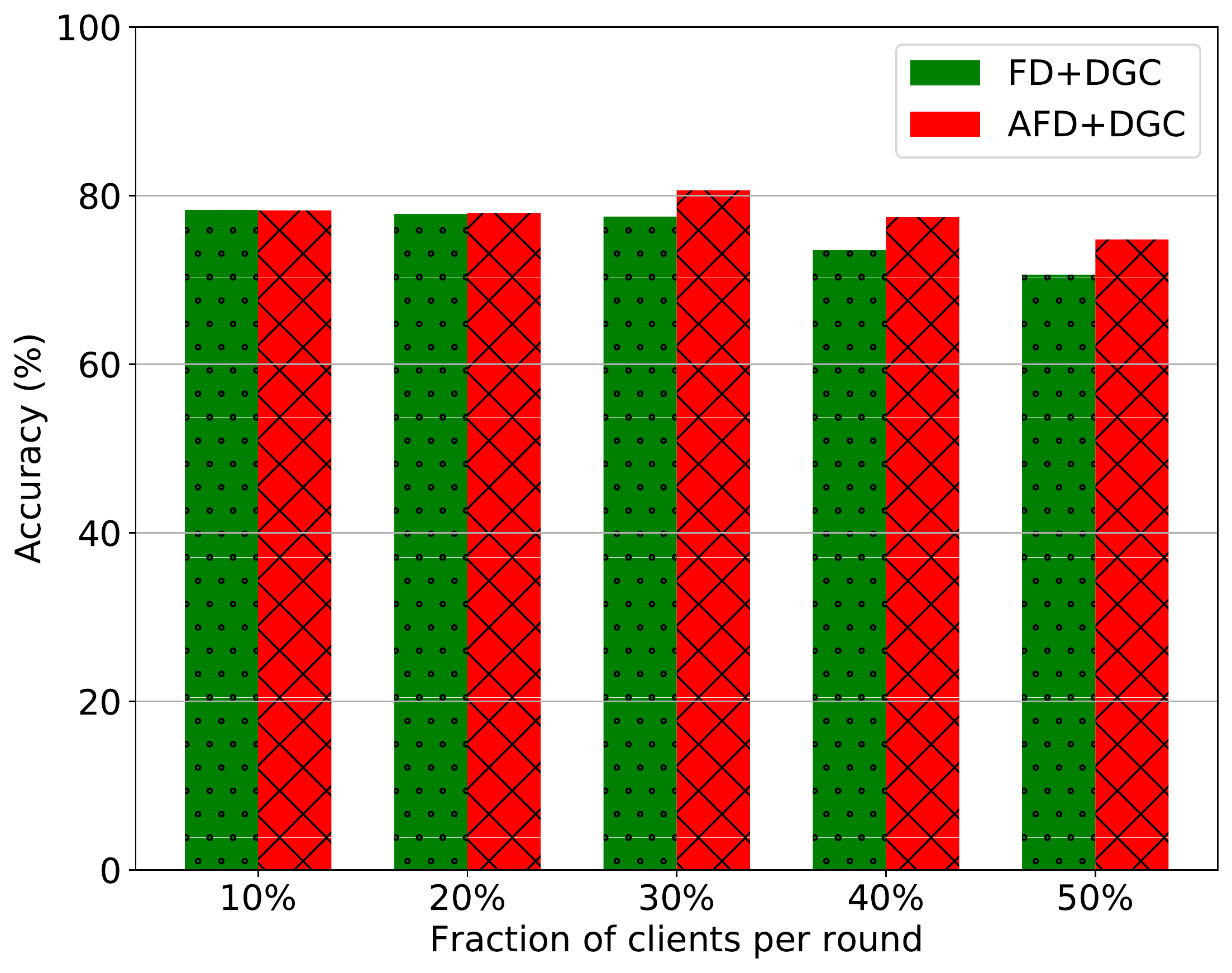}
\caption{Top-1 accuracy of Multi-Model AFD and FD when varying the fraction of clients per round in non-IID setting.}
\label{clients_rounds}
\end{figure}

The convergence time in Table~\ref{table1} represents the simulated wall-clock time to reach a pre-defined target accuracy. These results are obtained by simulating wireless links between the server and the clients based on the standard network speeds of Verizon 4G LTE wireless network which handles download speeds between 5 and 12 Mbps (Megabits per second) and upload speeds between 2 and 5 Mbps. All clients are supposed to experience the same network conditions. AFD manages to achieve speeds ups between 52$\times$ and 57$\times$, outperforming both DGC and FD + DGC, which only reduce the total convergence time by 31$\times$ to 44$\times$. Despite the fact that that AFD has a similar dropping ratio as FD, the generalization improvement systematically results in faster convergence, especially after the warm-up period to learn the activations score map.

\begin{table*}[t]
\centering
\caption{Accuracy and convergence time results on the IID LEAF datasets. For FEMNIST, Shakespeare and Sentiment140 datasets, the models are trained for 1000, 80 and 400 rounds, and the target accuracy is set to 82\%, 50\% and 83.5\% respectively.}
\label{table2}
\begin{tabular}{c|l|c|c|c}
\Xhline{1pt}
\multicolumn{2}{l|}{}  &  Accuracy & Convergence Time (min) & Speedup Ratio \\ \hline
\multirow{4}{*}{FEMNIST}  & No Compression  & 83.9\% $\pm$ 0.09\% &  3119.9 & 1$\times$ \\ 
                   & DGC & 83.6\% $\pm$ 0.27\% &  84.9 & 37$\times$ \\ 
                   & FD + DGC  & 84.1\% $\pm$ 0.72\% &  65.7 & 48$\times$ \\ 
                   & AFD + DGC  & \textbf{86.2\% $\pm$ 0.55\%} &  \textbf{58.1} & \textbf{53$\times$} \\ \hline
\multirow{4}{*}{Shakespeare}  & No Compression  & 52.2\% $\pm$ 0.18\% & 705.7 & 1$\times$ \\ 
                   & DGC & 50.8\% $\pm$ 0.85\% & 25.6 & 28$\times$ \\ 
                   & FD + DGC  & 50.9\% $\pm$ 0.72\% & 16.9 & 48$\times$ \\ 
                   & AFD + DGC  & \textbf{53.7\% $\pm$ 0.65\%} & \textbf{12.4} & \textbf{57$\times$} \\ \hline
\multirow{4}{*}{Sentiment140}  & No Compression  & 84.7\% $\pm$ 0.16\% & 2893.4 & 1$\times$ \\ 
                   & DGC & 84.5\% $\pm$ 0.77\% & 82.6 & 35$\times$ \\ 
                   & FD + DGC  & 84.5\% $\pm$ 0.39\% & 68.8 & 42$\times$ \\ 
                   & AFD + DGC  & \textbf{85.3\% $\pm$ 0.75\%} & \textbf{52.6} & \textbf{55$\times$} \\ 
\Xhline{1pt}
\end{tabular}
\end{table*}

As previously discussed, a small fraction of clients per round causes AFD to behave similarly to FD because of the highly inaccurate activations score map. That is evident by looking at Figure~\ref{clients_rounds}. Nevertheless, authors in~\cite{McMahan2017} suggest that increasing the amount of multi-client parallelism will not yield any advantages beyond a certain point, as also shown in Figure~\ref{clients_rounds}. We find that setting the fraction of clients at each round at 30-35\% leverages a good trade-off.  

Therefore, we tried a different mode of AFD, the Single-Model AFD, which only requires a single sub-model to be shared with the clients and, hence, the server needs to maintain a single activations score map. This map is updated at each training round based on the average loss function of the selected clients. In this case, the amount of multi-client parallelism cannot affect the AFD algorithm. However, Single-Model AFD is not effective when employed in non-IID environments.  That seems counter-intuitive at first sight, but the problem with non-IID environments is that the average loss function is calculated over different clients from one round to another. So, comparing two average losses from two different training rounds is not reliable because the global model can get biased and does perform the same with regards to all local datasets in the clients. Therefore, we consider IID setting in the remainder of this paper to make sure that the clients will have the same data distribution.

Figure~\ref{fig3} and Table~\ref{table2} summarize the performance results when we select the IID version of the datasets and using 10\% of the clients per round. For all three models, compression resulted in models with minor accuracy degradation under all compression schemes except for the Single-Model AFD. For convergence time, the later achieves speedups between 53$\times$ and 57$\times$ compared to only 42-48$\times$ for FD. Such observations prove the robustness of our technique and the high quality updates provided by the sub-model. 

\section{Conclusion}


In this paper, we propose and study Adaptive Federated Dropout, a practical method that trains high-quality models using a small communication footprint, as demonstrated by the experimental results on a variety of modes. We demonstrate the benefit of dynamically selecting subsets of the global modal on the trade-off between generalization and reducing communication costs. We empirically show that a combination of model compression and Adaptive Federated Dropout allows for up 57$\times$ speedup in convergence time compared to only up 44$\times$ for the state-of-the-art.

In future work, we plan to explore the effect of the dynamically selected sub-models on the fairness. Another future direction involves the personalization of clients' sub-models and freezing layers that reached the convergence stage.

\bibliography{references}
\end{document}